\documentclass[twoside,b5paper,10pt]{article}
\usepackage{AUTstyle}

\DeclareMathOperator{\EX}{\mathbb{E}}

\title{Importance of Environment Design in Reinforcement Learning: A Study of a Robotic Environment}
\author{M\'{o}nika Farsang \and Dr. Luca Szegletes}

\institution{Department of Automation and Applied Informatics \\
Budapest University of Technology and Economics}

\email{monika.farsang@edu.bme.hu \and luca.szegletes@aut.bme.hu}

\headerTitle{Importance of Environment\dots}
\headerAuthor{M\'{o}nika Farsang and Dr. Luca Szegletes}

\begin{document}
\makeAutStyleTitle

\begin{abstract}
An in-depth understanding of the particular environment is crucial in reinforcement learning (RL). To address this challenge, the decision-making process of a mobile collaborative robotic assistant modeled by the Markov decision process (MDP) framework is studied in this paper. The optimal state-action combinations of the MDP are calculated with the non-linear Bellman optimality equations. This system of equations can be solved with relative ease by the computational power of Wolfram Mathematica, where the obtained optimal action-values point to the optimal policy. Unlike other RL algorithms, this methodology does not approximate the optimal behavior, it gives the exact, explicit solution, which provides a strong foundation for our study. With this, we offer new insights into understanding the action selection mechanisms in RL by presenting various small modifications on the very same schema that lead to different optimal policies. 
\end{abstract}


\section{Introduction}
\label{sec:Introdu}

A number of approaches exist to find the optimal policy in reinforcement learning. Even with the best methods, there is a chance that we cannot achieve significant results. In this case, the explanation is often related to the insufficient reward design in the given setting. For this reason, shaping the reward function is increasingly becoming a vital factor beyond building efficient RL algorithms in order to obtain the desired behavior.

The motivation of this paper is to give novel visual representations of the importance of environment design in reinforcement learning by showing the connection between the reward function and the optimal behaviour. We also analyze the impact of the stochasticity on the action-selection mechanism. With them, we would like to show that small changes in the functions can result unexpectedly in new optimal policies. We study a mobile robotic assistant in a high-level abstraction of the MDP formulation to handle this challenging task. For finding the optimal policy, the Bellman optimality equations are computed directly in this paper.

Several methods have been proposed finding an optimal or near-optimal policy. Policy iteration ~\cite{howard:dp} and value iteration ~\cite{10.1287/mnsc.24.11.1127} are classical approaches to solve the Bellman equations. These dynamic programming (DP) techniques ~\cite{Bellman:1957} use the combination of policy evaluation and policy improvement. The former refers to the iterative computation of the value functions for a given policy. The latter means the computation of an improved policy for a given value function. Moreover, these techniques give the foundations of temporal difference (TD) learning proposed by Bertsekas and Tsitsiklis ~\cite{Bertsekas+Tsitsiklis:1996} and actor-critic algorithms by Konda and Borkar ~\cite{journals/siamco/KondaB99} as well. The advantage of TD methods over DP is that the reward function and transition probability distributions are not required ~\cite{Sutton1998}. 

Many RL approaches estimate the action-value functions from experience. If the agent performs each action in each state and stores the returns followed by them, the average return for each state-action pair will converge to the true values. This kind of technique is used by the Monte Carlo method, where the values are based on the average over many random samples from the environment ~\cite{10.1007/BF00114726}. In the case of complex environments, the values are estimated by parameterized functions, where the parameters are adjusted in order to match the observed returns ~\cite{Sutton1998}.

Watkins and Dayan developed Q-learning ~\cite{Watkins1992}, which is a simulation-based tabular method, where arrays or tables are used to store the approximations for each state-action pair. This widely used TD technique gives the basis of several algorithms. Later, it was combined with deep learning, resulting in Deep Q-Networks ~\cite{Mnih2013}.

Designing the proper reward function for a RL environment is a challenging task. The difficulty of having the right reward function can be presented by the credit assignment problem discussed by Minsky ~\cite{Minsky:1961:ire}, where the problem is the determination of the actions that lead to a specific outcome. As another aspect, shaping the reward function to accelerate learning often requires domain-expertise, demonstrated in ~\cite{Mataric94rewardfunctions}.

Experiments on sparse reward problems were conducted in recent years ~\cite{pathak2017curiositydriven, burda2018exploration,trott2019keeping}. Sparse reward problems occur even in simple tasks when the number of non-zero rewards is limited and encounters far away from each other in time. This usually leads to an agent behaving randomly for longer periods in the environment without gaining more experience. 

As another approach, MDPs with unknown reward functions were investigated via inverse reinforcement learning (IRL) by Ng and Russell ~\cite{Ng2000}. IRL showed to be useful for apprenticeship learning, where the goal is to learn the reward function of the expert from its behavior, which is a linear combination of known features ~\cite{10.1145/1015330.1015430}.

\section{Preliminaries}
A fundamental approach in reinforcement learning is formulating problems as MDPs ~\cite{Watkins1989}, which can be used in sequential decision making in stochastic environments. The formalism of an MDP includes sets of states $S$, actions $A$ and rewards $R$. The agent makes decisions $a_t \in A(s_t )$  based on the response of the environment at each timestep $t$. With the action $a_t$, it senses the next state $s_{t+1} \in S$ and receives reward $r_{t+1} \in R$. The transition probability distributions, denoted by $p$, are dependent only on the preceding state $s_t$ and action $a_t$ as shown in \equationname~\eqref{eq_p}. Consequently, these processes have the Markov property, which means that the future depends only on the present state and the past is not involved. 

\begin{equation}
\sum_{s' \in S} \sum_{r \in R} p(s', r\mid s, a)=1,\text{ for all }s \in S, a \in A(s)\label{eq_p}
\end{equation}

The particular way of selecting actions in each state is defined by the policy $\pi(a\mid s)$, $a \in A(s)$ for each $s\in S$. To measure the performance of the agent, the action-value function for policy $\pi$ is defined as can be seen in \equationname~\eqref{eq_qpi_exp}. It formulates the expected return when starting with the action $a$ in state $s$ and the present values of the future rewards are determined by the discount factor $0 \leq \gamma \leq 1$.

\begin{equation}
q_\pi(s,a) = \EX\left[{\sum_{k=0}^{\infty} \gamma^{k} r_{t+k+1} \mid  s,a}\right] \label{eq_qpi_exp}
\end{equation}

Computing the action-value functions can be done by using the Bellman equation in \equationname~\eqref{eq_qpi_bell}, which expresses the relation between the current action-value and its successor action-values ~\cite{Sutton1998}. It gives the sum of all possibilities taken action $a$ in state $s$ weighted by the transition probability $p$.

\begin{equation}
q_\pi(s,a) = \sum_{s',r}p(s',r\mid s,a) \left[ r+ \gamma \sum_{a'} \pi (a' \mid s') q_{\pi} (s', a')\right] \label{eq_qpi_bell}
\end{equation}

A policy is considered to be optimal if and only if $q_{\pi} (s,a) \geq q_{\pi'} (s,a)$, for all  $a \in A(s)$ $s \in S$, where $\pi$ is the current policy and  $\pi'$ is any other policy ~\cite{Sutton1998}. In other words, a policy is better than any other if its expected return in every state-action pair situation is greater than or equal to values with the same state-action combinations of other policies. This also means there could be multiple optimal policies, which are denoted by $\pi^{*}$. The solutions of the Bellman optimality equations, which is shown in \equationname~\eqref{eq_qpi_bellopt}, share the same value of optimal action-value functions $q_{*}(s,a)$.

\begin{equation}
q_*(s,a) = \sum_{s',r}p(s',r\mid s,a) \left[r+ \gamma \underset{a'}{\text{max}} q_{*} (s', a')\right] \label{eq_qpi_bellopt}
\end{equation}

\equationname~\eqref{eq_qpi_bellopt} formulate a non-linear equation for each state-action pair. The system of equations with the known environment behavior can be solved with relatively small effort. As a result, having the action-values leads to the optimal policy. Namely, the optimal action of the state maximizes the $q_{*}(s,a)$ value in the given state.

The abstract form of the MDPs provides a wide scale of applications. For example, low-level states and actions can be torque sensor readings and voltages applied to motors of a robotic joint. In contrast, high-level abstractions can be symbolic descriptions of an object, such as the place of the agent in a grid space with actions turning right or left.

\section{Methods}
This section first outlines the high-level abstraction form of the robotic environment examined in our study. In the second subsection, we present the used method of the optimal behavior calculation.

\subsection{Environment}\label{Env_subsevtion}
We now describe the experiment in which we analyze how the RL environment affects the optimal behavior. In our scenario, the task of a mobile collaborative robotic assistant is to collect empty soda cans in an office area. This problem is based on the recycling robot example by Sutton and Barto ~\cite{Sutton1998}, where the authors claimed that it was inspired by the mobile robot built by Connell ~\cite{10.5555/889083}. 

This environment is modeled by the MDP framework with high-level abstraction, where the state set $S=\{\text{High, Low}\}$ consists of the rechargeable battery levels of the robot. Its possible actions are searching actively for empty cans, waiting for someone to bring it one and recharging its own battery. 

Concerning the action sets  $A(s=\text{High})=\{\text{search\textsubscript{H}, wait\textsubscript{H}}\}$ and $A(s=\text{Low})=\{\text{search\textsubscript{L}, wait\textsubscript{L}, recharge\textsubscript{L}} \}$, the subscripts refer to the state where the action is performed. These sets are different in the two states because the recharging action in the high energy level state is unreasonable. The state-action combinations with the resulting states, their transition probabilities and rewards are described in \tablename~\ref{tab_dynamics}, where the notation is based on the work ~\cite{Sutton1998} with small modifications.

\begin{table}[htb]
\caption{Robotics environment dynamics}
\begin{center}
\begin{tabular}{|c|c|c|c|c|}
\hline
\multicolumn{3}{|c|}{\textbf{State-action pairs with next state}}&\multicolumn{2}{|c|}{\textbf{Environment}} \\
\hline
\textbf{state} & \textbf{action} & \textbf{next state} & \textbf{transition}& \textbf{reward} \\
\textbf{\textit{$s$}} & \textbf{\textit{$a$}} & \textbf{\textit{$s'$}} & \textbf{\textit{$p(s'\mid s,a)$}}& \textbf{\textit{$r(s,a, s')$}} \\
\hline
\multirow{3}{*}{High} &	\multirow{2}{*}{search\textsubscript{H}} &	High	& $\alpha$ &	$r_{search}$ \\
\cline{3-5}
 &	 &	Low	& $1-\alpha$ &	$r_{search}$ \\
 \cline{2-5}
 &	wait\textsubscript{H} &	High	& $1$ &	$r_{wait}$ \\
 \hline
\multirow{4}{*}{Low}&	\multirow{2}{*}{search\textsubscript{L}} &	High	& $1-\beta$ &	$r_{rescue}$ \\
\cline{3-5}
 &	 &	Low	& $\beta$ &	$r_{search}$ \\
 \cline{2-5}
 &	wait\textsubscript{L} &	Low	& $1$ &	$r_{wait}$ \\
 \cline{2-5}
 &	recharge\textsubscript{L} &	High	& $1$ &	$r_{recharge}$ \\

\hline
\end{tabular}
\label{tab_dynamics}
\end{center}
\end{table}

\begin{figure*}[!tb]
\captionsetup{justification=centering}
 \centering
 \includegraphics[width=\linewidth]{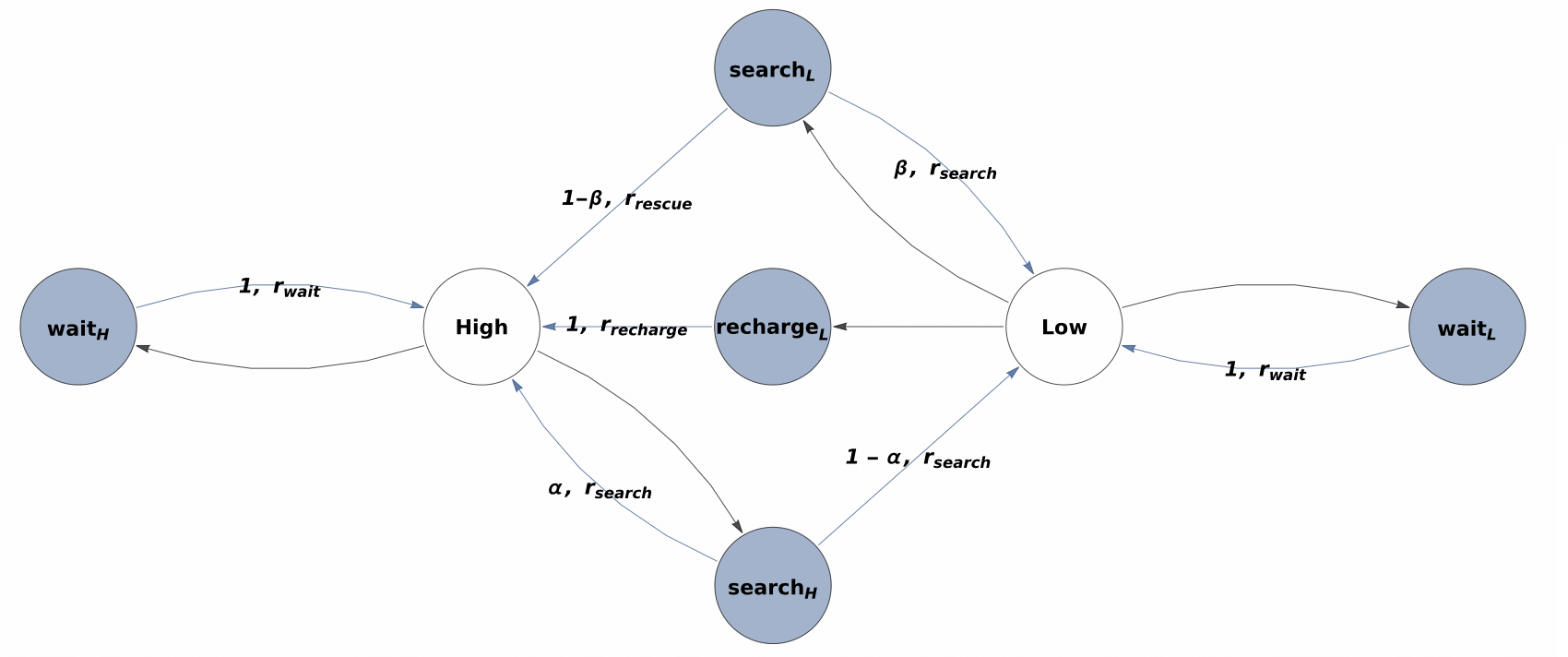}
 \caption{Transition graph of the MDP in the robotic assistant environment. The arrows between the selected action (blue) and the resulting state (white) nodes are labeled by the transition probability and the corresponding reward.}
 \label{fig_transition}
\end{figure*}

\figurename~\ref{fig_transition} presents the transition graph of the MDP. The states are represented by the white circles, where each state can be considered as the initial one. The corresponding actions are indicated by blue color. From the state nodes black edges connect to the action nodes indicating decision making. In reverse, blue edges labeled with the transition probability and reward pair come from the action nodes to the resulting state nodes.

The searching action in the high energy level state can result in two different outcomes. Meanwhile, the state stays the same with the probability $\alpha$, due to the consumed energy it switches to the low-power mode with $1-\alpha$. Similarly, when this action is performed in a low energy state, the next state will be the same with probability $\beta$ and the battery can become empty with the probability $1-\beta$, where the office personnel must rescue the mobile robot and bring it to the recharging station. 

The waiting action does not change the state but the number of collected cans highly depends on the people in the environment. For this reason, it is important to decide how desirable is to encourage this passive behavior. Finally, with choosing the recharging action type, the mobile robot assistant can recharge its battery when it is low, resulting in the high energy level state. 

\subsection{Calculation of the optimal actions}\label{Calc_subsection}

The aim of this paper is to give reliable results without the possible divergence of the chosen RL algorithm. Consequently, we calculate the action-value functions for the optimal policy based on the non-linear Bellman optimality equation as shown in \equationname~\eqref{eq_qpi_bellopt}. This technique was chosen because it is the most precise way to solve MDP problems.

With this classical method in the described environment, two equations correspond to the high energy level and three to the low energy level state with respect to the state-action pairs. To solve the system of equations, Wolfram Mathematica ~\cite{Inc.} was used thanks to its computational power.  

For the four different rewards ($r_{search}$, $r_{wait}$, $r_{recharge}$, $r_{rescue}$) and the transition probabilities ($\alpha$, $\beta$), a six-dimensional graph would show perfectly the impact of the parameters on the optimal actions in each state. Since it is not achievable, we present several 3D graphs and 2D plots in \figurename~\ref{fig:3d}, \figurename~\ref{fig:2d_rescue} and \ref{fig:2d_waiting}, which are easy to understand and give new insights on the optimal behavior. 

The not displayed parameters remain constant during each analysis and their specific values can be seen in the Appendix. To gain more experiments, our Wolfram Demonstration Project provides a much simpler but more interactive platform for the same problem at \url{https://demonstrations.wolfram.com/RecyclingRobotInReinforcementLearning/}. The Wolfram Notebook with the calculation and the graphs is publicly available at \url{https://www.wolframcloud.com/obj/farsangmonika95/Published/Importance of Environment Design in RL.nb}.

\section{Results and discussion}

This section begins by examining how the simultaneous alternation of two environmental parameters modify the optimal action. In this case, rewards of two actions and the two transition probabilities are analyzed. The second subsection investigates the impact of reward function by separately changing the rewards of the actions.

\subsection{Experiments of simultaneously changing the reward function of two actions and the transition probabilities}

\begin{figure}[!tb]
\captionsetup{justification=centering}
\centering
  \begin{minipage}{.5\linewidth}
\subfloat[\label{fig:3da}]{\includegraphics[width=\linewidth]{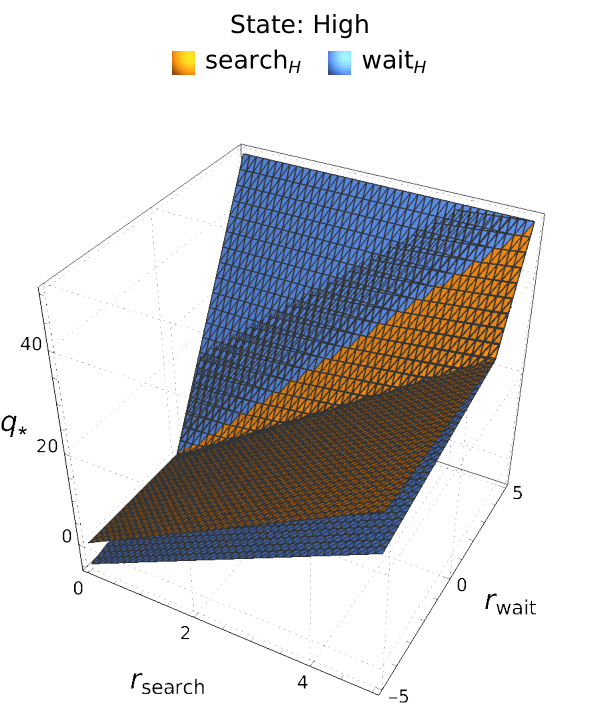}}
  \end{minipage}%
  \begin{minipage}{.5\linewidth}
    \subfloat[\label{fig:3db}]{\includegraphics[width=\linewidth]{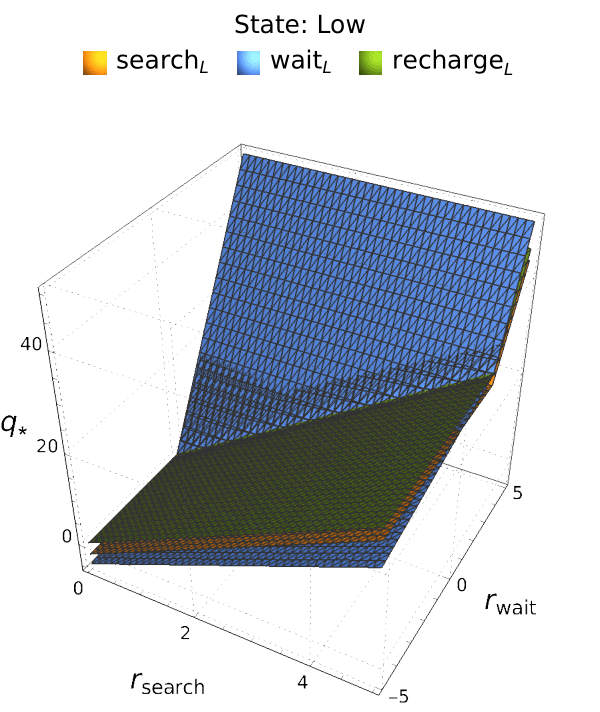}}
  \end{minipage}
\par\bigskip
    \begin{minipage}{.5\linewidth}
\subfloat[\label{fig:3dc}]{\includegraphics[width=\linewidth]{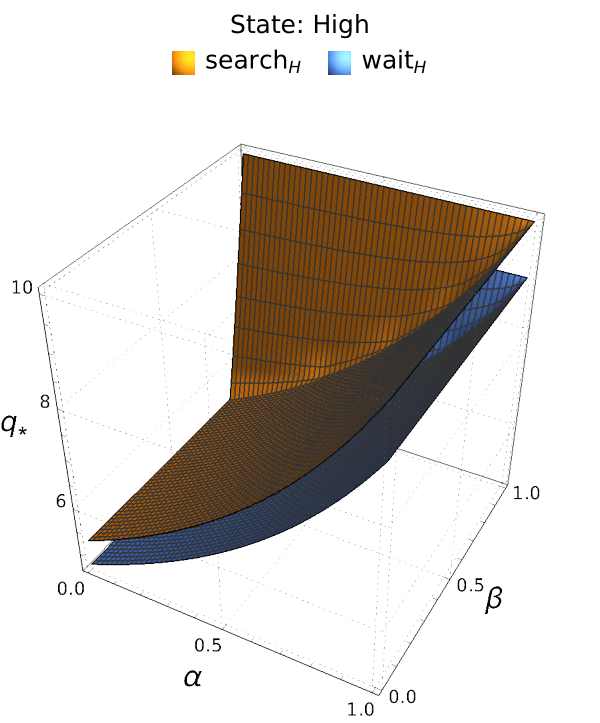}}
  \end{minipage}%
  \begin{minipage}{.5\linewidth}
    \subfloat[\label{fig:3dd}]{\includegraphics[width=\linewidth]{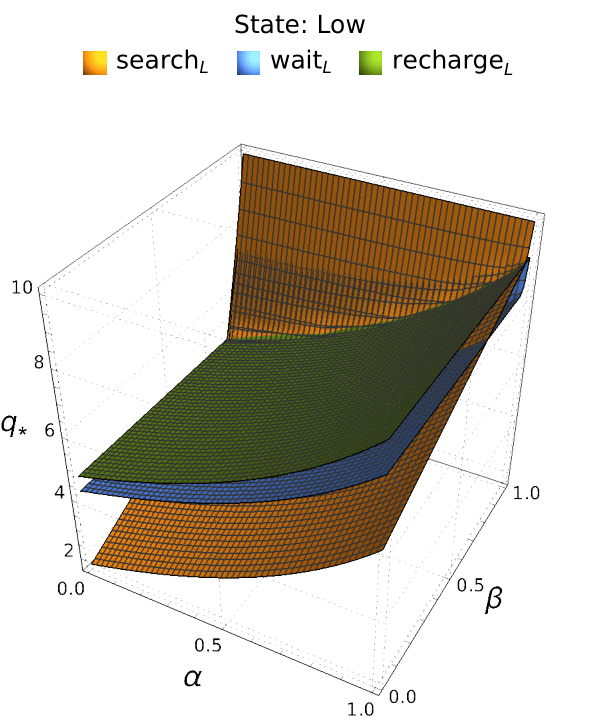}}
  \end{minipage}
  \caption{Examination of the reward function and the transition probabilities. (a) Impact of the searching and waiting reward on the high energy level state. (b) Same inspection as (a), but on the low energy level state.  (c) Influence of the transition probabilities on the high-level battery. (d) Same analysis as (c), but on the low-level battery.}
  \label{fig:3d}
\end{figure}

In our first set of experiments, we investigate the impact of two environmental attributes on the optimal policy. \figurename~\ref{fig:3d}\subref{fig:3da} and \ref{fig:3d}\subref{fig:3db} present the searching and waiting reward connection to the optimal action-value with the high-level and the low-level battery respectively. 

The high energy level state has only two possible actions, which are searching and waiting, and exactly their reward shaping is analyzed here. In this case, it is self-evident that the action with a higher reward becomes optimal in this state. Meanwhile, it is not obvious how the optimal action changes in the low-power mode. The searching reward has an indirect impact on the recharging action and the searching action never becomes optimal in this scenery.  Although, it is worth mentioning, that with other fixed parameter setting it has the possibility to get the searching action optimal with the low-level battery as well. For example, as can be seen in \figurename~\ref{fig:3d}\subref{fig:3dd}, the modified transition probabilities can lead to a completely different optimal policy.

In \figurename~\ref{fig:3d}\subref{fig:3dc} and \ref{fig:3d}\subref{fig:3dd}, we analyze the transition probabilities of the environment. They determine the probability of the next state after performing the searching action in high energy level state ($\alpha$) and low energy level state ($\beta$). Interestingly, we found that the modification of these values does not change the optimal action with the high-level battery, where the plane of searching action remains above the waiting action-values in the whole range. However, the optimal action of the low-power mode can be immensely influenced simply by the transition probabilities. Here, the recharging action is optimal in most of the cases, but if the transition probability $\beta$ has a high enough value, the searching action exceeds that and become the optimal choice.

\subsection{Experiments of separately changing the reward function of two actions}

\begin{figure}[htb]
\captionsetup{justification=centering}
     \begin{minipage}{.5\linewidth}
        \centering
        \subfloat[\label{fig:2d_rescue_a}]{\includegraphics[width=\linewidth]{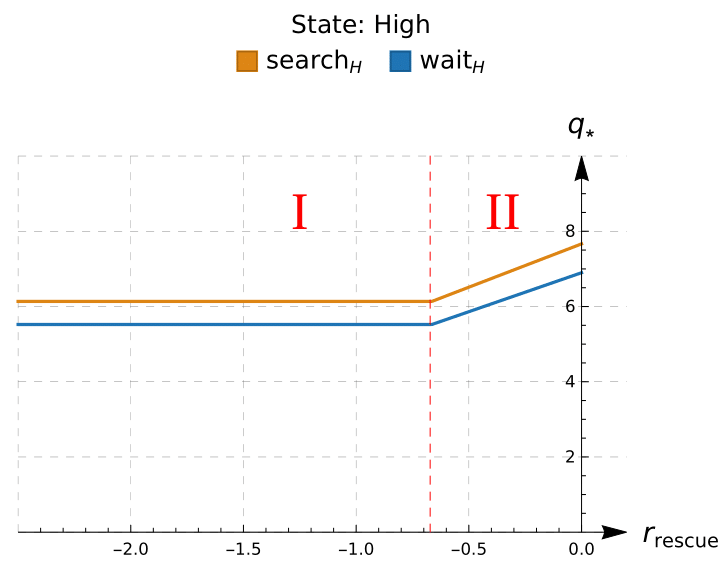}}
        \end{minipage}%
        \begin{minipage}{.5\linewidth}
        \subfloat[\label{fig:2d_rescue_b}]{\includegraphics[width=\linewidth]{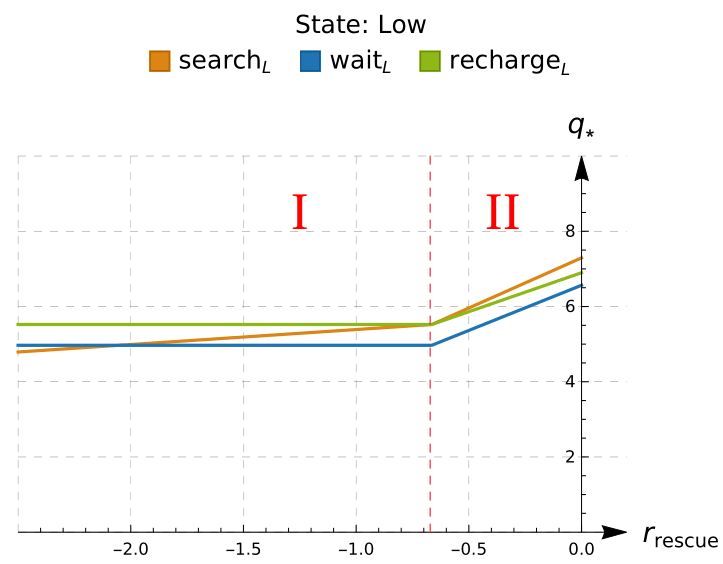}}
        \end{minipage}
        \caption{Changes in optimal actions by modifying the rescue reward in the high and low energy level states. (a) Effect of the rescue reward on the state with high-level battery. (b) Same analysis as (a), but with low-level battery.  }
        \label{fig:2d_rescue}
\end{figure}

\begin{figure}[htb]
\captionsetup{justification=centering}
        \begin{minipage}{.5\linewidth}
        \subfloat[\label{fig:2d_wait_a}]{\includegraphics[width=\linewidth]{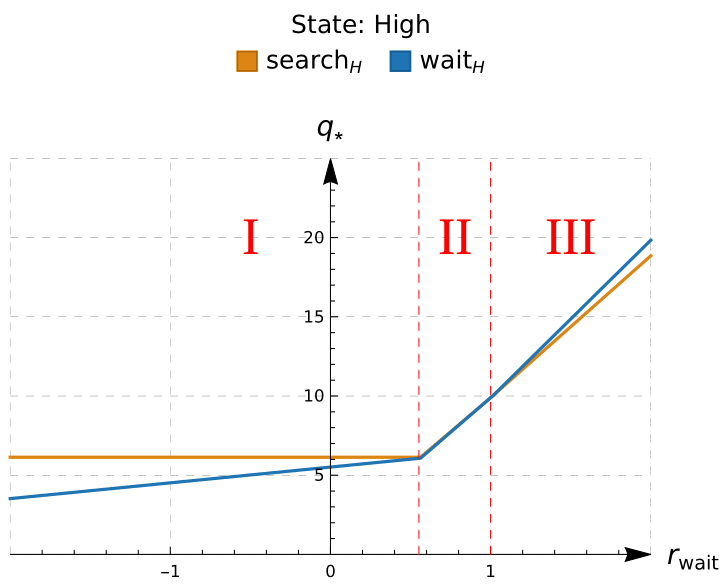}}
        \end{minipage}%
         \begin{minipage}{.5\linewidth}
        \subfloat[\label{fig:2d_wait_b}]{\includegraphics[width=\linewidth]{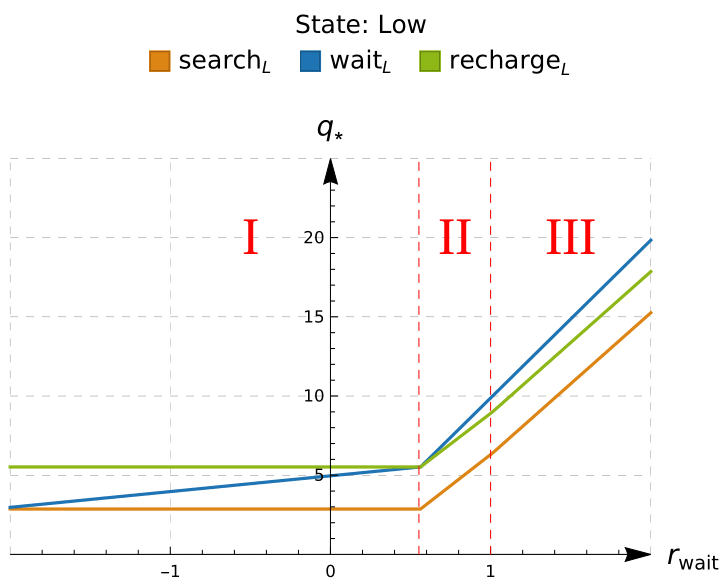}}
        \end{minipage}
        
        \caption{Impact of the waiting reward on the high and low energy level states. (a) Analyzing how the waiting reward affects the state with high level battery. (b) Same examination as (a), but the low-power state.}
        \label{fig:2d_waiting}
\end{figure}

The second set of our experiments contain simpler two-dimensional plots, where we investigated how the minor changes in one specific parameter affect the chosen optimal actions. 
\figurename~\ref{fig:2d_rescue} and \ref{fig:2d_waiting} illustrate the effect of the rescuing and the waiting reward separately, where the red vertical lines present the boundary when the optimal policy changes in one of the two states.

\paragraph{Changing the rescuing reward} \hfill
\vfill
The rescuing reward occurs, if the robot starts searching with the low-level battery. As can be seen in \figurename~\ref{fig:2d_rescue}\subref{fig:2d_rescue_a} and \ref{fig:2d_rescue}\subref{fig:2d_rescue_b}, positive rewards are not displayed because this kind of behavior should be avoided. This penalty does influence the action-values in the high and the low energy level states as well. On one hand, it does not change the optimal searching action in the high energy level state throughout the analysis. On the other hand, as the rescuing reward increases, three different scenarios can be observed in the low energy level state. 
\vfill
\vfill
In the first period, the recharging action is the optimal choice. At one point around -0.67, the searching action-value becomes identical with the value of the recharging action, illustrated by the red vertical line. In this case, both actions are considered to be optimal. 

With steadily increasing the rescue reward, the searching action surpasses the recharging action in the second phase and becomes the only optimal action in the low-level battery state. The switching point of the optimal behavior can be unexpected without this kind of analysis.

\paragraph{Changing the waiting reward} \hfill
\vfill
\figurename~\ref{fig:2d_waiting}\subref{fig:2d_wait_a} and \ref{fig:2d_waiting}\subref{fig:2d_wait_b} illustrate the influence of the waiting reward, which is received when the mobile robot is inactive and constantly waits for the office personnel to bring it cans. In this case, negative and positive rewards are considered, because it depends on the expected activity of the robotic assistant. There are five possible optimal behaviors based on the value of this reward, represented by the labeled sections and their boundaries. 
\vfill
\vfill
In the first phase, the searching and recharging actions are optimal in the high and the low energy level state respectively. As the value of the waiting action increases in the low-power mode, it reaches the recharging value and both become optimal at around 0.55, presented in \figurename~\ref{fig:2d_waiting}\subref{fig:2d_wait_b}. 

In the second part, the searching action in the high and the waiting action in the low energy level state are the best choices. Later, the action-value of the waiting action rapidly increases and both actions become optimal when the waiting reward equals 1 in the high-level battery state, represented by the second vertical red line in \figurename~\ref{fig:2d_waiting}\subref{fig:2d_wait_a}. 

With the further increment of the waiting reward, the corresponding waiting action is the only optimal decision in both states in the third section. This differs hugely from the optimal actions in the first part, even though there is only a small growth in the reward of the waiting.

This last part of our experiments demonstrates that the same reward can influence differently the states caused by the transition probabilities. Moreover, it shows that the modification of the reward function in a relatively small range has a huge effect on the optimal behavior in both states, resulting even in multiple optimal actions.

\section{Conclusion}
In this paper, we have investigated the impact of the environment design on reinforcement learning problems. For this, we selected an MDP robotic task, where several adjustments were presented in the environment. The results of this study support the idea that even small differences in the chosen reward function and transition probabilities can affect the outcome. 

Our research has highlighted the importance of these aspects by the direct computation of the non-linear Bellman optimality equations and novel visualization techniques by showing the connection between the optimal action-values and the environmental parameters. We suggest this method of analysis in each environment where it is applicable to understand the underlying cause of the learned behavior.

\section*{Acknowledgments}
 { \small The author would like to express her thanks to Dr. Luca Szegletes for her support as a scientific advisor.
This work has been supported by the Budapest University of Technology and Economics.}

\section*{Appendix}

\begin{table}[htb]
\caption{Environment parameters in the different experiments}
\begin{center}
\begin{tabular}{|c|c|c|c|c|}
\hline
\multirow{4}{*}{\textbf{Parameter}}&\multicolumn{4}{|c|}{\textbf{Values in the experiments}} \\
\cline{2-5}
 & \textbf{Exp. 1.}  & \textbf{Exp. 2.}  & \textbf{Exp. 3.}  & \textbf{Exp. 4.} \\
  & \figurename~\ref{fig:3d}\subref{fig:3da},  & \figurename~\ref{fig:3d}\subref{fig:3dc},  & \figurename~\ref{fig:2d_rescue}\subref{fig:2d_rescue_a},  & \figurename~\ref{fig:2d_waiting}\subref{fig:2d_wait_a}, \\
    & \ref{fig:3d}\subref{fig:3db} & \ref{fig:3d}\subref{fig:3dd}  & \ref{fig:2d_rescue}\subref{fig:2d_rescue_b}  & \ref{fig:2d_waiting}\subref{fig:2d_wait_b} \\
\hline
$r_{rescue}$ & -3 & -3 & -2.5$\dots$0 & -3  \\
\hline
$r_{recharge}$ & 0 & 0 & 0 & 0  \\
\hline
$r_{wait}$ & -5$\dots$5 & 0  & 0  & -2$\dots$2 \\
\hline
$r_{search}$ & 0$\dots$5 & 1 & 1 & 1   \\
\hline
$\alpha$ & 0.3 & 0$\dots$1 & 0.3 & 0.3   \\
\hline
$\beta$ & 0.1 & 0$\dots$1 & 0.6 & 0.1    \\
\hline
$\gamma$ & 0.9 & 0.9 & 0.9 & 0.9 \\
\hline
\end{tabular}
\label{tab_parameters}
\end{center}
\end{table}

\makeAutBib{bibtexdatabase}

\end{document}